\documentclass[conference]{IEEEtran}
\usepackage{times}

\usepackage[numbers]{natbib}
\usepackage{multicol}
\usepackage[bookmarks=true]{hyperref}
\usepackage{mathtools}
\usepackage{amsmath,amssymb,amsfonts}
\usepackage{xcolor}
\usepackage[noadjust,sort]{cite}
\usepackage[font=footnotesize]{caption}
\usepackage{subcaption}
\usepackage{booktabs}
\usepackage{algorithm}
\usepackage{algpseudocode}
\usepackage{graphicx}
\usepackage{lipsum} 
\usepackage{xcolor}
\usepackage{color}
\usepackage{hyperref}
\usepackage{etoolbox}
\usepackage{textcomp}
\usepackage{multirow}

\algnewcommand\algorithmicswitch{\textbf{switch}}
\algnewcommand\algorithmiccase{\textbf{case}}
\algdef{SE}[SWITCH]{Switch}{EndSwitch}[1]{\algorithmicswitch\ #1\ \algorithmicdo}{\algorithmicend\ \algorithmicswitch}%
\algdef{SE}[CASE]{Case}{EndCase}[1]{\algorithmiccase\ #1}{\algorithmicend\ \algorithmiccase}%
\algtext*{EndSwitch}%
\algtext*{EndCase}%

\usepackage[normalem]{ulem} 

\renewcommand{\paragraph}[1]{\vspace{0em}\noindent\textbf{#1}.}

\usepackage{color}
\usepackage{xcolor}
\definecolor{turquoise}{cmyk}{0.65,0,0.1,0.3}
\definecolor{purple}{rgb}{0.65,0,0.65}
\definecolor{dark_green}{rgb}{0, 0.5, 0}
\definecolor{orange}{rgb}{0.8, 0.6, 0.2}
\definecolor{red}{rgb}{0.8, 0.2, 0.2}
\definecolor{darkred}{rgb}{0.6, 0.1, 0.05}
\definecolor{blueish}{rgb}{0.0, 0.3, .6}
\definecolor{light_gray}{rgb}{0.7, 0.7, .7}
\definecolor{pink}{rgb}{1, 0, 1}
\definecolor{greyblue}{rgb}{0.25, 0.25, 1}
\definecolor{tab_blue}{HTML}{1f77b4}
\definecolor{tab_orange}{HTML}{ff7f0e}
\definecolor{LightRed}{rgb}{0.99,0.89,0.89}

\definecolor{mesh_misty_rose}{HTML}{e6aaa3}
\definecolor{mesh_yellow}{HTML}{ffba00}

\definecolor{our_red}{rgb}{0.99,0.89,0.89}
\definecolor{our_blue}{HTML}{1f77b4}
\definecolor{our_orange}{HTML}{ff7f0e}

\newcommand{\SupplementaryMaterial}{{\color{darkred} supplementary materials}}

\usepackage{amsmath,amsfonts,bm}

\def\eqref#1{equation~\ref{#1}}

\def\1{\bm{1}}

\def\va{{\bm{a}}}

\DeclareMathAlphabet{\mathsfit}{\encodingdefault}{\sfdefault}{m}{sl}
\SetMathAlphabet{\mathsfit}{bold}{\encodingdefault}{\sfdefault}{bx}{n}

\newcommand{\wrt}{\textit{w.r.t.}~}

\begin{document}

\title{\LARGE
\textbf{SAGE}: Bridging \textbf{{S}}emantic and \textbf{{A}}ctionable Parts for\\
\textbf{{Ge}}neralizable Manipulation of Articulated Objects}


\author{\authorblockN{Haoran Geng$^{*1,2}$, Songlin Wei$^{*2,3}$, Congyue Deng$^{1}$, Bokui Shen$^{1}$, He Wang$^{\dagger2,3}$, Leonidas Guibas$^{\dagger1}$}
\authorblockA{Department of Computer Science, Stanford University${^1}$ \\
CFCS, School of Computer Science, Peking University${^2}$ \\
Beijing Academy of Artificial Intelligence${^3}$ \\
$^*$ Equal contribution. $^\dagger$ Equal advising. \\ 
{Corresponding author: \texttt{hewang@pku.edu.cn}}
}
}


%



\twocolumn[{%
\renewcommand\twocolumn[1][]{#1}%
\maketitle

\begin{center}
    \centering
    \vspace{-2em}
    \captionsetup{type=figure}
    \includegraphics[width=0.95\linewidth]{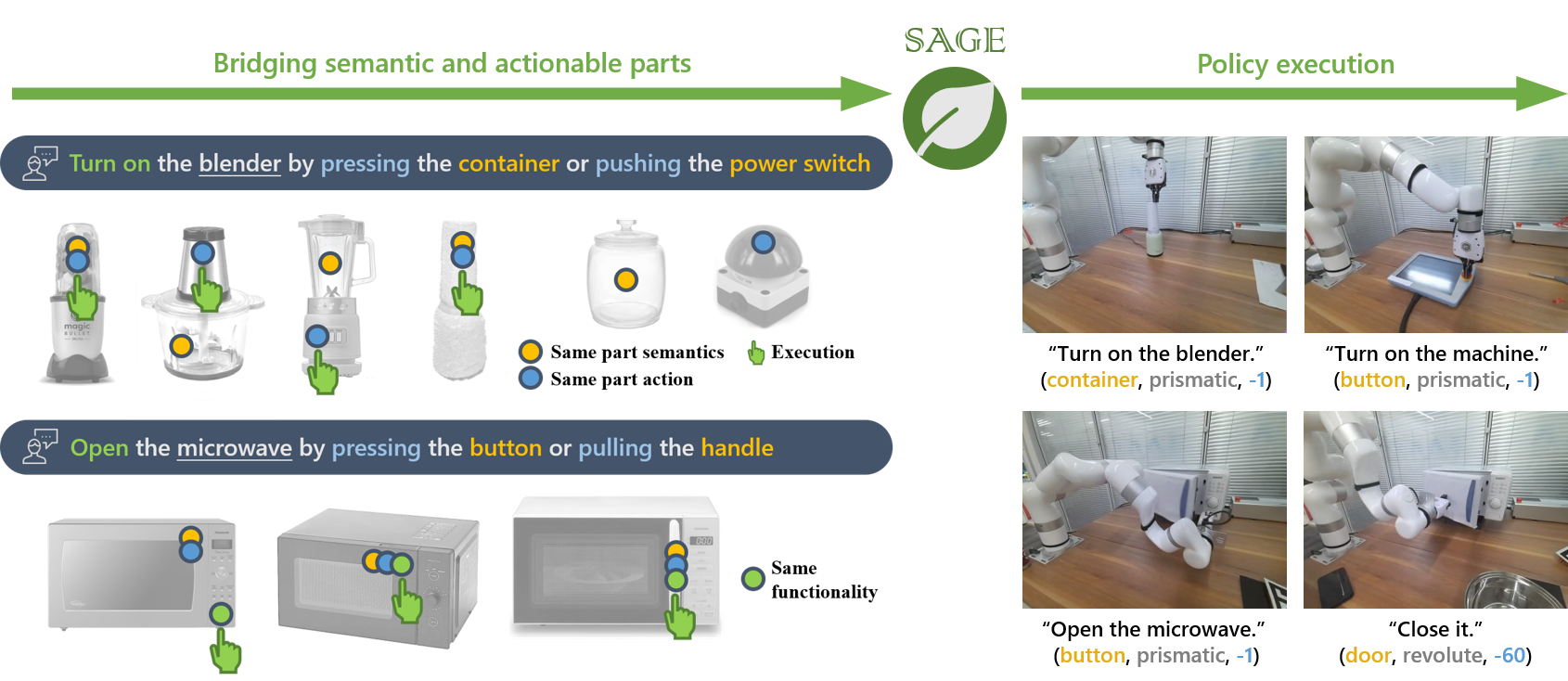}
    \captionof{figure}{\textbf{Overview.} We present SAGE, a framework bridging the understanding of semantic and actionable parts for generalizable manipulation of articulated objects. \textbf{Left:} We give two examples of human instructions illustrating the concept of part semantics, part functionalities, and corresponding actions. \textbf{Right:} examples of real-world tasks and our results. Our framework can tackle diverse manipulation tasks on various articulated objects.}
    \label{fig:teaser}
\end{center}
}]


\begin{abstract}

To interact with daily-life articulated objects of diverse structures and functionalities, understanding the object parts plays a central role in both user instruction comprehension and task execution.
However, the possible discordance between the semantic meaning and physics functionalities of the parts poses a challenge for designing a general system.
To address this problem, we propose SAGE, a novel framework that bridges semantic and actionable parts of articulated objects to achieve generalizable manipulation under natural language instructions.
More concretely, given an articulated object, we first observe all the semantic parts on it, conditioned on which an instruction interpreter proposes possible action programs that concretize the natural language instruction. Then, a part-grounding module maps the semantic parts into so-called Generalizable Actionable Parts (GAParts), which inherently carry information about part motion. End-effector trajectories are predicted on the GAParts, which, together with the action program, form an executable policy. Additionally, an interactive feedback module is incorporated to respond to failures, which closes the loop and increases the robustness of the overall framework.
Key to the success of our framework is the joint proposal and knowledge fusion between a large vision-language model (VLM) and a small domain-specific model for both context comprehension and part perception, with the former providing general intuitions and the latter serving as expert facts.
Both simulation and real-robot experiments show our effectiveness in handling a large variety of articulated objects with diverse language-instructed goals.

\end{abstract}
\section{Introduction}

From furniture to home appliances, articulated object prevails in our daily lives.
While being long studied in the robotics literature, their diverse object structures, functionalities, and manipulation goals in real-world scenarios remain challenging for most robot systems.
The design of a general system that accommodates such object diversities falls into the problem of generalizable object manipulation, which involves the transfer of manipulation skills across object shapes \citep{mu2021maniskill, gu2023maniskill2}, object categories\citep{geng2022gapartnet, geng2023partmanip, xu2023unidexgrasp, wan2023unidexgrasp++}, or even tasks \citep{yu2021metaworld}.
Specifically, a line of works has focused on learning generalizable manipulation skills of articulated objects \citep{mu2021maniskill, gu2023maniskill2, geng2022gapartnet, geng2023partmanip, gong2023arnold, geng2022end}.
However, even these works are still limited to a few common object categories, leaving aside the problem of how to model the resemblance between seemingly irrelevant articulated objects, from a simple cabinet to a multifunctional blender with intricate mechanical structures and electricity-powered functionalities.

In this work, we approach this problem by separately modeling the cross-category part commonality in their semantics and actions. Taking the blender in Fig. \ref{fig:teaser} top left as an example, to mince the food ingredients in it, one needs to press its container from the top to turn it on.
The part you press on is semantically a ``container'', but considering your interaction with it, its action resembles a ``button'' which can be pressed and trigger other mechanical motions (such as the blade rotating). To exert the functionality of this object, both part semantics and actions should be well understood.
This also fits into the two different types of affordances in Gibson's theory of distributed cognition \citep{zhang2006distributed}: physical affordance and cognitive affordance, which have different emphases in human cognition of interactable objects. Physical affordance focuses on physical structures such as the joint conditions and part motions of articulated objects; cognitive affordances, on the other hand, are provided by cultural conventions and involve a high-level semantic understanding of the object.

Following this inspiration, we introduce SAGE, a framework for generalizable manipulation of articulated objects under natural language instructions by bridging semantic and actionable parts (Fig. \ref{fig:teaser} right).
Our key insight is that large Visual-Language Models (VLMs) possess general knowledge of part semantics, while small domain-specific models present higher accuracy in predicting part actions, which can serve as ``expert facts''.
Different from prior works that separately assign VLMs and small models to different sub-tasks\citep{huang2023voxposer}, we fuse their predictions in both context comprehension and part perception, which achieves a good balance of generality and exactness.

Concretely, given a manipulation goal specified by natural language and an RGBD image as visual observation, an instruction interpreter first translates the language instruction into programmatic action representation. These action programs are composed of so-called action units, which are represented by 3-tuples of object semantic parts, joint types, and state changes.
We then convert the action programs defined on the object semantic parts into executable policies through a part grounding module that maps the semantic parts into so-called Generalizable Actionable Parts (GAParts), which is a cross-category definition of parts according to their actionabilities.
From the detected GAParts, we can generate physically plausible part actions indicated by the pre-obtained action programs.
Finally, we also introduce an interactive feedback module that actively responds to failed action steps and adjusts the overall policy accordingly, which enables the framework to act robustly under environmental ambiguities or failure.

We demonstrate the effectiveness of our method both in simulation environments and on real robots. 
Specifically, we also assess the crucial components of language-guided articulated-object manipulation: scene descriptions, part perceptions, and manipulation policies. Quantitatively, our method surpasses all baseline approaches. This success is attributed to our blend of a comprehensive general-purpose model and a precise domain specialist, endowing our method with both stellar performance and robust generalization capabilities. Qualitatively, we further illustrate the proficiency of our method through its performance on several demanding tasks.

To summarize, our key contributions are:
\begin{itemize}
    \item We bridge the notion of semantic and actionable parts to model the commonality across articulated objects of highly diverse structures and functionalities.
    \item We build a robot system for generalizable manipulation of articulated objects under language instructions.
    \item We design a knowledge fusion mechanism between VLMs and small domain-specific models to incorporate expert facts into general perception and comprehension.
    \item We demonstrate our strong generalizability on a variety of different objects under diverse language instructions in both simulation environments and on real robots.
\end{itemize}
\section{Related Work}

\paragraph{Articulated-object manipulation}
Articulated object manipulation presents significant challenges due to diverse object geometries and physical characteristics. Although benchmarks by \citep{urakami2019doorgym, mu2021maniskill} have been introduced, their scope remains limited. While methods like \citep{peterson2000high, jain2009pulling, burget2013whole} have delved into motion planning, visual affordance learning has also gained attention \citep{mo2021where2act, wu2022vatmart, wang2022adaafford, zhao2022dualafford, geng2022end}. 
Other works \citep{eisner2022flowbot3d, xu2022universal} have proposed special representations for manipulation, though they are tailored mainly for suction grippers. 
On the other hand, \citep{geng2022gapartnet, geng2023partmanip} introduce generalizable parts as a novel representation but are limited to single-part interaction and can not handle natural instructions.


\paragraph{Generalizable object manipulation} Achieving generalization in robot applications is essential yet challenging. Various works \citep{fang2020graspnet, sundermeyer2021contact, gou2021rgb, gao2021kpam, xu2021adagrasp} employ supervised learning with motion planning to tackle generalizable tasks such as grasping. Nevertheless, these techniques often necessitate task-specific architectures and may falter in intricate manipulations. Though reinforcement learning holds promise for complex challenges \citep{rajeswaran2017dapg, akkaya2019solving}, the lack of generalizability remains unresolved \citep{kirk2021generalization_survey,ghosh2021generalization}. The ManiSkill benchmark \citep{mu2021maniskill} pioneers category-level object manipulation in simulations, while \citep{shen2022learning} utilizes imitation learning with RL-trained demonstrations. However, in both cases, the generalization ability is still unclear.
\citep{geng2022gapartnet, geng2023partmanip} first tackles generalizable manipulation in a cross-category manner, but is still limited to predefined part classes. \citep{xu2023unidexgrasp, wan2023unidexgrasp++} have universal generalization ability but only for grasping tasks. In summary, Generalizable object manipulation is still extremely challenging.

\paragraph{Robot control with large language models}
Incorporating pre-trained language models into robot systems has been gaining tremendous attention recently. A notable segment of these works concentrates on the grounding of language to manipulation skills, essential for executing long-horizon instructions~\citep{jang2022bc,stepputtis2020language,mees2022matters,codevilla2018driving,shao2021concept2robot,akakzia2021grounding,gong2023arnold,goyal2021zero,shridhar2022cliport,hu2019hierarchical, you2023make}. A hierarchical approach, seen in another subset of studies, hinges on a two-fold process: establishing a skill set and subsequently strategizing over it using large language models (LLMs) \citep{brohan2023can,huang2022inner,singh2022progprompt,huang2022language,lin2023text2motion,jiang2022vima}. However, this method is limited by the huge data collection effort, is still constrained to the predefined skill sets, and lacks visual understanding. In contrast, LLMs are employed to spawn code for low-level skills through APIs, though predicting these skills' continuous parameters remains challenging \citep{codeaspolicies2022}.
Some employ a way-point-based method for efficiency, sidestepping low-level robot actions in favor of trajectory hand poses~\citep{shridhar2023perceiver, gong2023arnold}. But they also require lots of data and have no detailed planning strategy. VoxPoser
\citep{huang2023voxposer} performs motion planning based on affordance and obstacle maps generated by LLM, but still lacks visual understanding.
For articulated objects with diverse geometries, intricate structures, and physical constraints, language-guided manipulation remains a challenging and underexplored problem.


\section{Problem Formulation}

\subsection{Semantic and Actionable Parts}
As shown in Fig~\ref{fig:teaser}, we characterize a \textit{semantic part} based on its human-defined identity or its functional role in articulated objects. During the interaction, for a \textit{actionable part}, we adopt the concept of "Generalizable Actionable Part (GAPart)" as formulated by GAPartNet~\citep{geng2022gapartnet}. GAPart categorizes parts across different objects based on their actionability, where parts within the same GAPart category exhibit similar geometry and functionality. For instance, a button designed for pressing and a knob meant for rotation would not fall under the same GAPart category, even if they appear similar. Our approach emphasizes the importance of understanding parts at \textit{both semantic and action} levels for generalized manipulation of articulated objects.



\subsection{Problem Definition and Framework Overview}

\begin{figure}
    \centering
    \includegraphics[width=.95\linewidth]{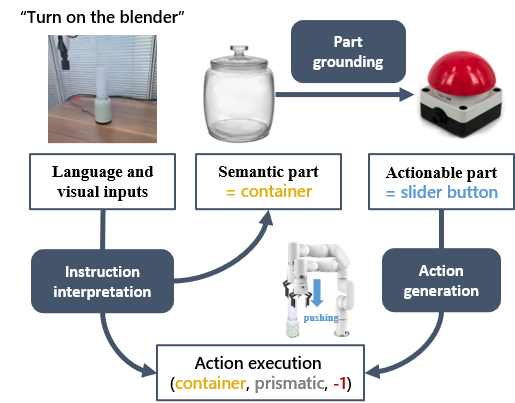}
    \caption{\textbf{Framework overview.} Given a natural language input conditioned on visual observation, we first obtain the semantic parts and interpret the language instructions on them. Then we link the semantic parts to actionable parts through a part grounding module, on which executable actions can be predicted.}
    \label{fig:framework_overview}
\end{figure}

\noindent \textbf{Problem definition.} Our task focuses on open-vocabulary manipulation of articulated objects. The input consists of a human language instruction and a single-view RGB-D image. A fixed base robot arm is required to follow human instructions and manipulate the articulated object within the scene to achieve the goal specified in the instructions.

\noindent \textbf{Framework overview (Fig. \ref{fig:framework_overview}).} We propose SAGE, a comprehensive framework designed for open-vocabulary articulated object manipulation in the wild. By combining the strength of the general-purpose Visual Language Model (VLM) and a domain-specific 3D part perception and manipulation model, SAGE is uniquely capable of understanding object parts on both semantic and action levels, facilitating more effective and adaptable manipulation in real-world scenarios.

In the SAGE system, we employ the classic \textit{perception-decision-execution-feedback} loop to generalizable articulate object manipulation:
More specifically, (1) we first use GPT-4V to process the input RGB image to obtain a scene description that contains task-related information, such as \textit{semantic parts}, part interaction possibilities, and object states. We further detect more \textit{actionable parts} with GAPartNet and fuse both outputs to obtain the final part-aware scene description. (2) Then, we again prompt GPT-4V to comprehend the scene description together with the human instruction to act like a global planner. It outputs action programs step by step and makes decisions based on execution feedback. (3) The action programs are executed with a motion planner which controls the robot arm to follow a predefined trajectory grounded on the actionable part. (4) Finally, we interactively detect the part and object state changes and feedback on the updates to the global planner.

\section{Method}
\label{sec:method}

\subsection{Part-aware Scene Perception}
\label{sec:method:scene_percep}

In daily-life scenarios, human language instructions could be diverse and even ambiguous. To facilitate large language models to truly understand human intention and plan accordingly, it is beneficial to add comprehensive scene descriptions to the prompt text inspired by the success of Chain-of-thought \citep{wei2023chainofthought} prompting.

\begin{figure}
    \centering
    \includegraphics[width=1\linewidth]{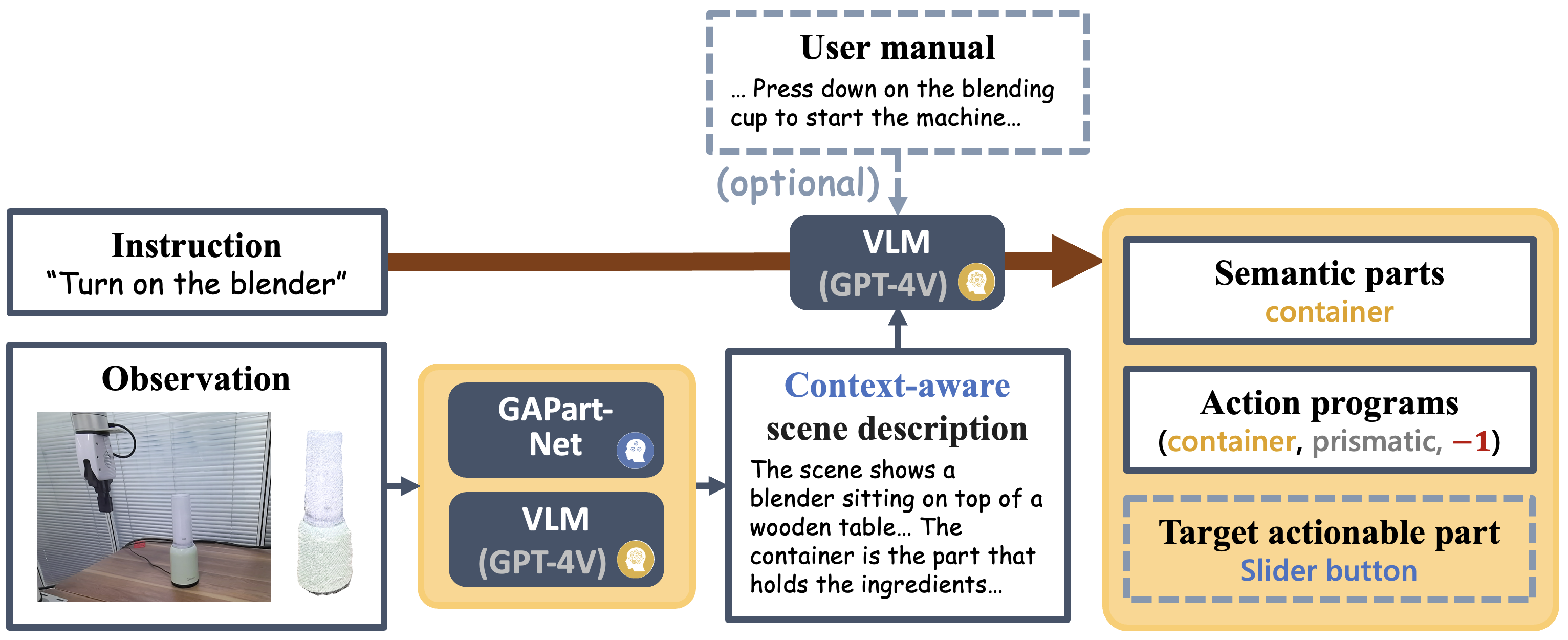}
    \caption{
    \textbf{Context-aware instruction interpretation.} With instruction and observation (RGBD image) as input, the interpreter first generates a scene description with VLM and GAPartNet\citep{geng2022gapartnet}. Then LLM(GPT-4) takes both instruction and scene description as input and generates semantic parts and action programs. Optionally, we can input a specific user manual, and LLM will generate a target actionable part; see Sec.\ref{sec:grounding} for more details about the actionable part.
    }
    \label{fig:pipe_interpretation}
\end{figure}

\begin{figure}
    \centering
    \includegraphics[width=0.95\linewidth]{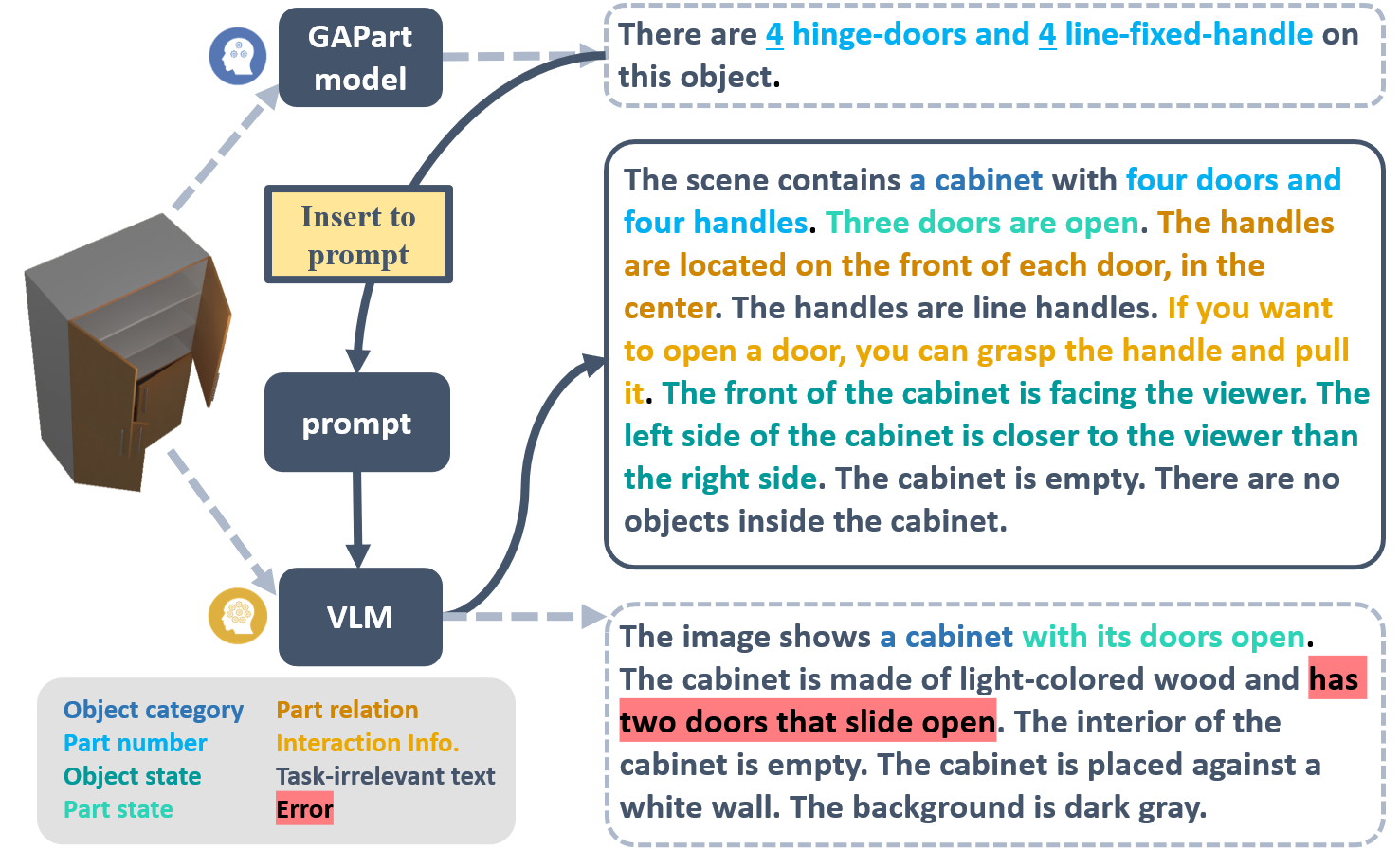}
    \caption{
    \textbf{Scene description.} To better help with action generation, the scene description should contain object information, part information, and some interaction-related information. We use expert GAPart model\citep{gadre2021act} to generate some expert descriptions as part of the VLM prompt and then generate the scene descriptions, which works well and absorbs the advantages of both models.
    }
    \label{fig:pipe_scene_descri}
    \vspace{-0.5cm}
\end{figure}

\paragraph{Scene Description}
Chain-of-thought\citep{wei2023chainofthought} is proven useful for prompting large models. Here, we also notice that, compared to outputting an action program (See Sec. \ref{ssec:action_program}) directly using a Visual-Language Model, it is better to first generate a scene description $\tilde{D}_{scene}$ containing task-related information (such as parts, object states, interaction possibilities).
However, we observe that VLM provides rich context but lacks accuracy in part-related perception as illustrated in Fig. \ref{fig:pipe_scene_descri}. This example illustrates that employing the Visual-Language Model directly can result in the omission of crucial information related to parts and tasks, or may even introduce factual inaccuracies.

\paragraph{Actionable Part Detection}
%
%
We deploy the 3D GAPart Model \citep{geng2022gapartnet} to enhance part perception. This model interprets the partial RGBD point cloud $X$ to yield its set of 3D part proposals, which are masked point clouds denoted as $\mathcal{M}^{3D} = \{l_j, M_j^{3D}\}_{j=1}^n$, where $l$ is the actionable part label, ${M}$ is the masked partial point cloud, and $j$ indexes the detected actionable parts from $1$ to $n$. We reframe the actionable parts with a template sentence as shown in Fig. \ref{fig:pipe_scene_descri}, eg., \textit{There are 4 hinge-door and 4 line-fixed-handle on the object}. The sentences that contain actionable parts are then appended to obtain the part-enhanced scene description ${D}_{scene}$. We empirically found this significantly boosts the performance of perception. More details are given in the experiments.

\subsection{Instruction Interpretation and Global Planner}
\label{language_method}
Given the part-aware scene description ${D}_{scene}$, we here again employ GPT-4V \citep{openai2023gpt4} as the instruction interpreter to translate natural-language instruction $\tilde{I}$ into executable actions that will be sent to the downstream execution module. 
We devise a representation named \textit{action unit} to encapsulate the output actions.

\paragraph{Action Units}
\label{ssec:action_program}
We define the most basic manipulation on one single part of articulated objects as an ``action unit'', which can be represented by a 3-tuple $\va = (\tilde{p}, j, \Delta s)$ with part name $\tilde{p}$, joint type $j$, and the change of part state $\Delta s$.
The part name $\tilde{p}$ is a noun in natural language.
$j$ refers to the joint directly linked to part $\tilde{p}$ and indicates whether it is a revolute or a prismatic joint; $\Delta s$ is the state change of joint $j$ under the action, which is either an angle $\pm\theta$ if $j$ is revolute or a relative translation $\pm t$ \wrt the part bounding box if $j$ is prismatic.
The final output of the interpreter should be in one of the following formats:
\begin{itemize}
    \item a single action unit,
    \item an unordered union of multiple action units,
    \item an ordered list of multiple action units,
\end{itemize}
or be a finite combination of the three formats.
Following the convention of programming languages (PL), this can be viewed as a typing system with expressions:
\begin{equation*}
    \va ~|~ \texttt{Union}\{\va,\va\} ~|~ \texttt{List}[\va,\va]
\end{equation*}
Here the \texttt{Union} expression is for non-deterministic policy generation when multiple action units can reach the same goal, e.g., both pulling the door and pressing a button result in the microwave door being opened.
\texttt{List} is for sequential action generation when no single-step solution exists, e.g., to open a door, a knob must be first rotated to unlock the door.

\paragraph{Global Planner} 
\label{ssec:global_planner}
Throughout the interaction, tracking of both the target gripper and part states is maintained by a global planner, where details are given in Algorithm \ref{algo:planning_update}. If significant discrepancies arise, the planner has the discretion to select from one of four states: "continue", "transition to the next step", "halt and replan", or "success". For instance, if the gripper is set to rotate 60 degrees along a joint, yet the door only opens to 15 degrees, the LLM planner would opt to "halt and re-plan". This interaction tracking model ensures that the LLM remains cognizant of developments during interactions, permitting strategy adjustments and, where necessary, recovery from unforeseen setbacks.

\subsection{Part Grounding and Execution}
\label{sec:grounding}

\begin{figure}[t]
    \centering
    \includegraphics[width=.85\linewidth]{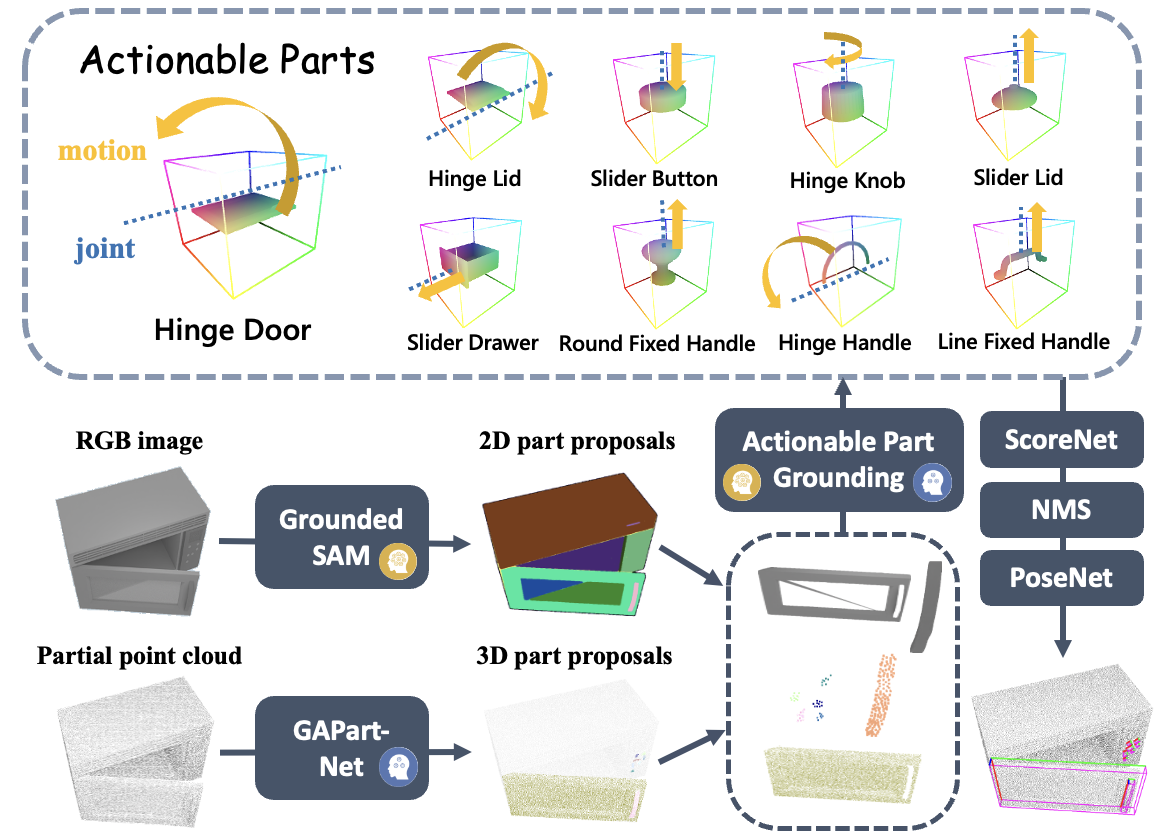}
    \caption{\textbf{Part grounding.} With the observation as input, our method ensembles 2D proposals from GroundedSAM and 3D proposals from GAPartNet. then the proposals are fed to the ground as an actionable part. Leveraging ScoreNet, NMS, and PoseNet, we then present the perception results. Notice that: (1) For the part perception evaluation benchmark, we directly employ SAM\citep{kirillov2023segany}. However, within our manipulation pipeline, we utilize GroundedSAM, which also considers the semantic part as input.
    (2) As highlighted in Fig.~\ref{fig:pipe_interpretation}, if the LLM produces a target actionable part, the grounding process is bypassed.
    }
    \label{fig:part_grounding}
    \vspace{-0.5cm}
\end{figure}




An execution module is implemented to act accordingly to the action unit outputted by the global planner.
Like any language-guided manipulation task, we first need to ground the action onto the corresponding physical actionable part.

\paragraph{Actionable part grounding} 
To this end, We first collect a dataset of actionable part features and part labels $\{{F}^\text{dino}_\text{part}, l_\text{part}\}$. 
This dataset contains the pre-defined interaction policies of each typical GAPart. The policies are a series of pre-defined end-effector trajectories to complete a certain motion for a specific part. It serves as our domain-specific expert which is the key to the success of our generalizable system.
In particular,
we extract a good image feature map $F^\text{dino}$ from DINOv2 \citep{oquab2023dinov2}. Then, for the target actionable part denoted as $p_j$, we use its mask to get the part feature $F^\text{dino}_j$.
We first do max-pooling to obtain the part feature and run KNN algorithm to find the grounded actionable part label denoted as $l_j$
\begin{align*}
    F^\text{dino}_j &= \text{MaxPooling}(F^\text{dino} [M_j]) \\
    l_j &= \text{KNN}(\{{F}^\text{dino}_\text{part}, l_\text{part}\}, F_j^\text{dino}).
\end{align*}


\paragraph{Trajectory Generation}
Once we have grounded the semantic part to the actionable GAPart, we can generate executable manipulations on this part. We first estimate the part pose $P_j$ as defined in GAPartNet \citep{geng2022gapartnet}. We also compute the joint state (part axis and position) and plausible motion direction based on the joint type (prismatic or revolute).
Then we generate the part actions according to these estimations. We first predict an initial gripper pose as the primary action. It is followed by a motion induced from a pre-determined strategy defined in GAPartNet \citep{geng2022gapartnet} according to the part poses and joint states. For example, to open a door with a revolute joint, the starting position can be on the door's rim or handle, with the trajectory being a circular arc oriented along the door's hinge.



\subsection{Interactive Feedback}

\begin{algorithm} 
    \caption{Global Planner} 
    \label{alg4}
    \hspace*{\algorithmicindent} \textbf{Input} Gripper target and current state $\Delta g_t$,  $\delta g_t$ and part target and estimated movement $\Delta s_t$, $s_t$. 
    \begin{algorithmic}[1]
        \State $prompt$ = $\text{template}(\Delta g_t, \delta g_t, \Delta s_t, s_t)$ 
        \State $result$ = call VLM with $prompt$
        \Switch{$result$}
            \Case{$\texttt{"continue"}$}
                \State{Continue the current strategy}
            \EndCase
            \Case{$\texttt{"transition to next step"}$}
                \State{Finish the current action tuple and transition to the next action.}
            \EndCase
            \Case($\texttt{"halt and replan"}$)
                \State{Stop the current execution and replan} \Comment{Some errors happened. }
            \EndCase
            \Case($\texttt{"success"}$)
                \State{Success and terminate. }
            \EndCase
        \EndSwitch
    \end{algorithmic}
    \label{algo:planning_update}
\end{algorithm}
\begin{algorithm}
    \caption{Interactive perception}
    \label{alg3} 
    \hspace*{\algorithmicindent} \textbf{Input} initial and current part point cloud $X_\text{part}^0, X_\text{part}^t$  \\
    \hspace*{\algorithmicindent} \textbf{Output} joint axis $j_t$ and movement $s_t$ 
    \begin{algorithmic}[1]
        \State $t_t, R_t = \text{RANSAC}(\text{Umeyama}(X_\text{part}^0, X_\text{part}^t))$
        \State $j_t, s_t = \text{GeometryInference}(R_t, t_t)$
    \end{algorithmic}
    \label{algo:interactive_perception}
\end{algorithm}

Up to now, we have only utilized a single initial observation $I^0_{rgbd}$ for generating open-loop interactions. We now introduce a mechanism to further leverage the observations acquired during the interaction process, which can update the perception results and adjust the manipulations accordingly. Toward this goal, we introduce a two-section feedback mechanism during interaction.

\noindent\textbf{Interactive perception.} 
It should be noted that occlusion and estimation errors may arise during the perception of the first observation. To address these challenges, we propose a model that leverages interactive observations to enhance operation (see Algo.~\ref{algo:interactive_perception}). Consider two distinct observations, 
$I_{rgbd}^{0}$ and $I_{rgbd}^{t}$, captured during an interaction. A part's motion can be detected between these two observations. We subsequently employ the mask of the moved part to deduce the actual joint state during the interaction. Applying RANSAC \citep{fischler1981random} for outlier removal and Umeyama algorithm \citep{umeyama1991least}, we estimate the rotation and translation between the two frames of the part's point cloud $X_\text{part}^0, X_\text{part}^t$. This computation subsequently allows us to ascertain the joint state
$j_t$ and the current part movement states $s_t$.
Finally, we send the joint state and part movement state back to the global planner (see Sec. \ref{ssec:global_planner}, thereby completing the main loop.

\section{Experiments}

\begin{figure}[t]
    \centering
    \includegraphics[width=.85\linewidth]{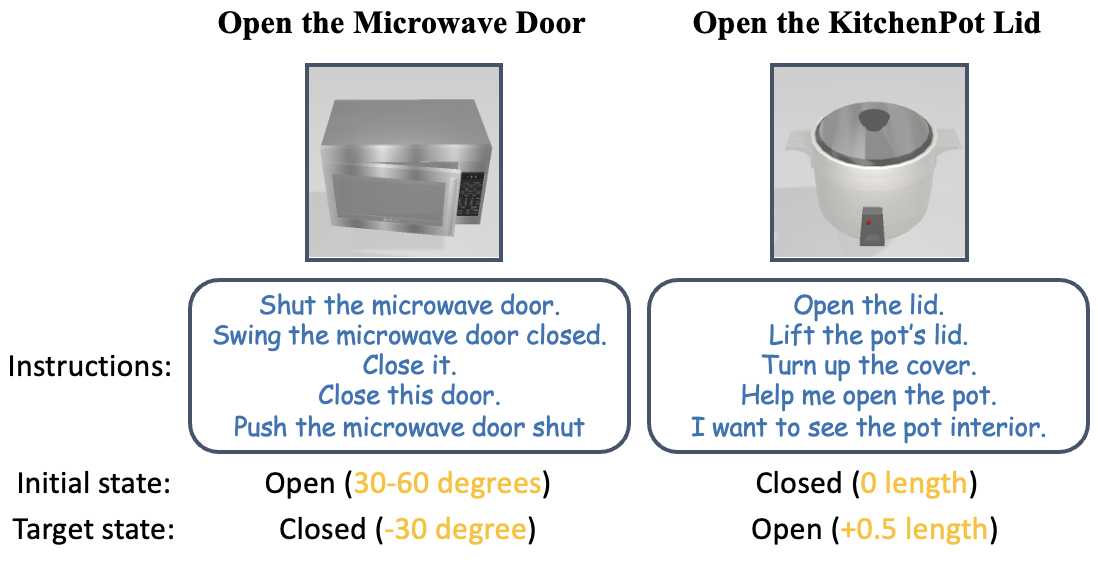}
    \caption{\textbf{Task examples for experiments in simulation.}}
    \label{fig:bench_demo}
    \vspace{-0.5cm}
\end{figure}

\begin{table*}[t]
    \centering
    \resizebox{2.\columnwidth}{!}{
\begin{tabular}{l|ccc|ccc|ccc|c|c|c}
\toprule
Category & \multicolumn{3}{c|}{Microwave} & \multicolumn{3}{c|}{StorageFurniture} & \multicolumn{3}{c|}{Cabinet} & KitchenPot & Remote & Blender \\ \midrule
Task ID & 1 & 2 & 3 & 4 & 5 & 6 & 7 & 8 & 9 & 10 & 11 & 12 \\ \midrule
VoxPoser$^*$\citep{huang2023voxposer} & - & - & 13.0 & - & - & 15.0 & - & - & 14.2 & - & - & - \\
GAPartNet$^*$\citep{geng2022gapartnet} & 87.5 & 75.5 & 58.6 & 53.3 & 76.9 & 81.3 & 50.0 & 66.3 & 73.8 & - & - & - \\
Ours & \textbf{98.0} & \textbf{96.0} &\textbf{94.1} & \textbf{83.3} & \textbf{80.0} &\textbf{95.0} & \textbf{79.3} & \textbf{89.6} & \textbf{83.1} &\textbf{85.7} & \textbf{60.7}& \textbf{42.9}\\ \bottomrule
\end{tabular}
}
\caption{\textbf{Success rates (\%) under language instructions.} Comparison between our method with VoxPoser \citep{huang2023voxposer} and GAPartNet\citep{geng2022gapartnet}. We evaluate 12 tasks with 6 different articulated objects. "-" means the task is not implemented by the baseline. GAPartNet$^*$ shares the same execution policy as our method, while VoxPoser$^*$ is adapted from the unofficial version of the authors.}
\vspace{-0.5cm}
\label{tab:main_results}
\end{table*}

\begin{table}[t]
    \centering
    \setlength{\tabcolsep}{3pt}
    \resizebox{\columnwidth}{!}{
    \begin{tabular}{l|cccc}
    \toprule
    Category & \#Task & \#Init. state & \#Tgt. state & \#Instruct. \\
    \midrule
    Microwave & 100 & 3 & 2 & 5 \\
    StorageFurniture & 40 & 3 & 2 & 5 \\
    Cabinet & 40 & 3 & 2 & 5 \\
    KitchenPot & 20 & 1 & 1 & 5 \\
    Remote & 18 & 1 & 1 & 5 \\
    Blender & 20 & 1 & 1 & 5 \\
    \bottomrule
    \end{tabular}}
    \caption{\textbf{Benchmark statistics.} For each object category, we created tasks that were randomly initialized.}
    \label{tab:bench_stat}
\end{table}

\begin{table}[t]
\centering
\resizebox{0.7\columnwidth}{!}{
\begin{tabular}{l|l|l}
\toprule
\multicolumn{1}{c|}{\multirow{2}{*}{Perception}} & Scene Description          & \textless{}1s \\
\multicolumn{1}{c|}{}                            & Part Detection             & 13s           \\ \midrule
\multirow{2}{*}{Decision}                         & Instruction Interpretation & \textless{}10s \\
                                                  & Global Planninng           & \textless{}1s \\ \midrule
\multirow{2}{*}{Execution}                        & Part Grounding             & \textless{}5s \\
                                                  & Motion Planning            & 45s           \\ \midrule
Feedback                                          & Interactive Perception     & \textless{}1s \\ \bottomrule
\end{tabular}
}
\caption{\textbf{Average runtime of each module.}}
\label{tab:runtime}
\end{table}

\subsection{Simulation Experiments}

\paragraph{Setup}
We conducted our simulations using the SAPIEN environment \citep{xiang2020sapien} and designed 12 language-guided articulated object manipulation tasks. An example task is given in Fig. \ref{fig:bench_demo}. A comprehensive breakdown of these tasks, along with detailed statistics, is presented in Table \ref{tab:main_results} \ref{tab:bench_stat}.
For each category of \textit{Microwave}, \textit{StorageFurniture} and \textit{Cabinate}, we devised 3 tasks including opening from the initial open and close state and closing from the initial open state. The remaining tasks for \textit{KitchenPot}, \textit{Remote} and \textit{Blender} are \textit{Open the lid},  \textit{Press the button} and \textit{Turn on the blender} respectively.
For each task, we carefully curated more than 5 distinct objects sourced from the GAPartNet dataset \citep{geng2022gapartnet} that were well-suited for the respective actions. For example, we selected 5 different models of \textit{Microwaves} from GAPartNet, chosen specifically for tasks like \textit{pulling the door open} and \textit{pushing the door closed}. Likewise, we chose 20 different \textit{StorageFurniture} objects and 10 \textit{Blenders}, each tailored to their corresponding tasks.
To assess the robustness of our part interaction module, we conducted over 20 trials for each task. During each trial, we introduced randomization in both camera position and initial joint states, ensuring the variability of scenarios. Specifically, for the task \textit{Close the door}, the initial positions of the articulated object doors were randomized within the range of $(30, 60)$ degrees.
To facilitate fair comparisons, we leveraged pre-trained weights from GAPartNet to identify actionable parts within the scene. Subsequently, we applied the same motion generation policy as our pipeline.

\paragraph{Results}
The results of our experiments are summarized in Table \ref{tab:main_results}, showcasing the superior performance of our methods across nearly all tasks.
The key drivers of our success can be attributed to our enhanced perception capabilities, which benefit from the fusion of the specialized model from GAPartNet and the generalist model of LLM. The average runtime breakdown of each module in the SAGE system is shown in Table. \ref{tab:runtime}.

\begin{figure}[t]
    \centering
    \includegraphics[width=1\linewidth]{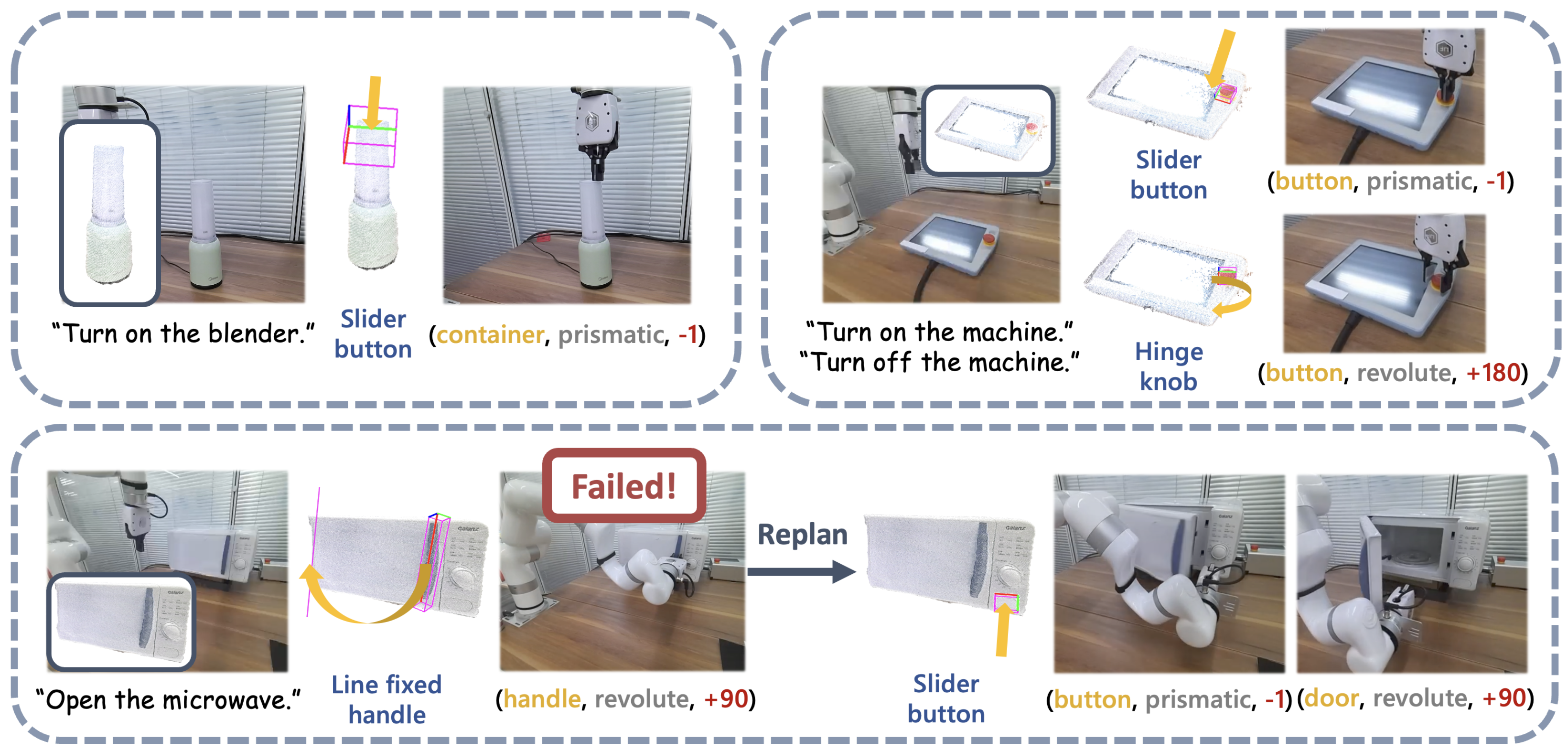}
    \caption{\textbf{Explanations of real-world results.}
    \textbf{Top:} "turn on the blender" and "turn on/off the machine". Our method accurately understands the part semantics and actions that are not aligned.
    \textbf{Bottom:} "Open the microwave". The mechanical structures of the microwave prevent the robot from directly pulling open the door but instead require the button to be pressed, resulting in a failure. However, our interactive feedback model can detect the failure and recognize that it should try pressing the button instead and subsequently complete the task.}
    \label{fig:exp_real_detail}
\end{figure}

\begin{figure*}
  \includegraphics[width=\textwidth]{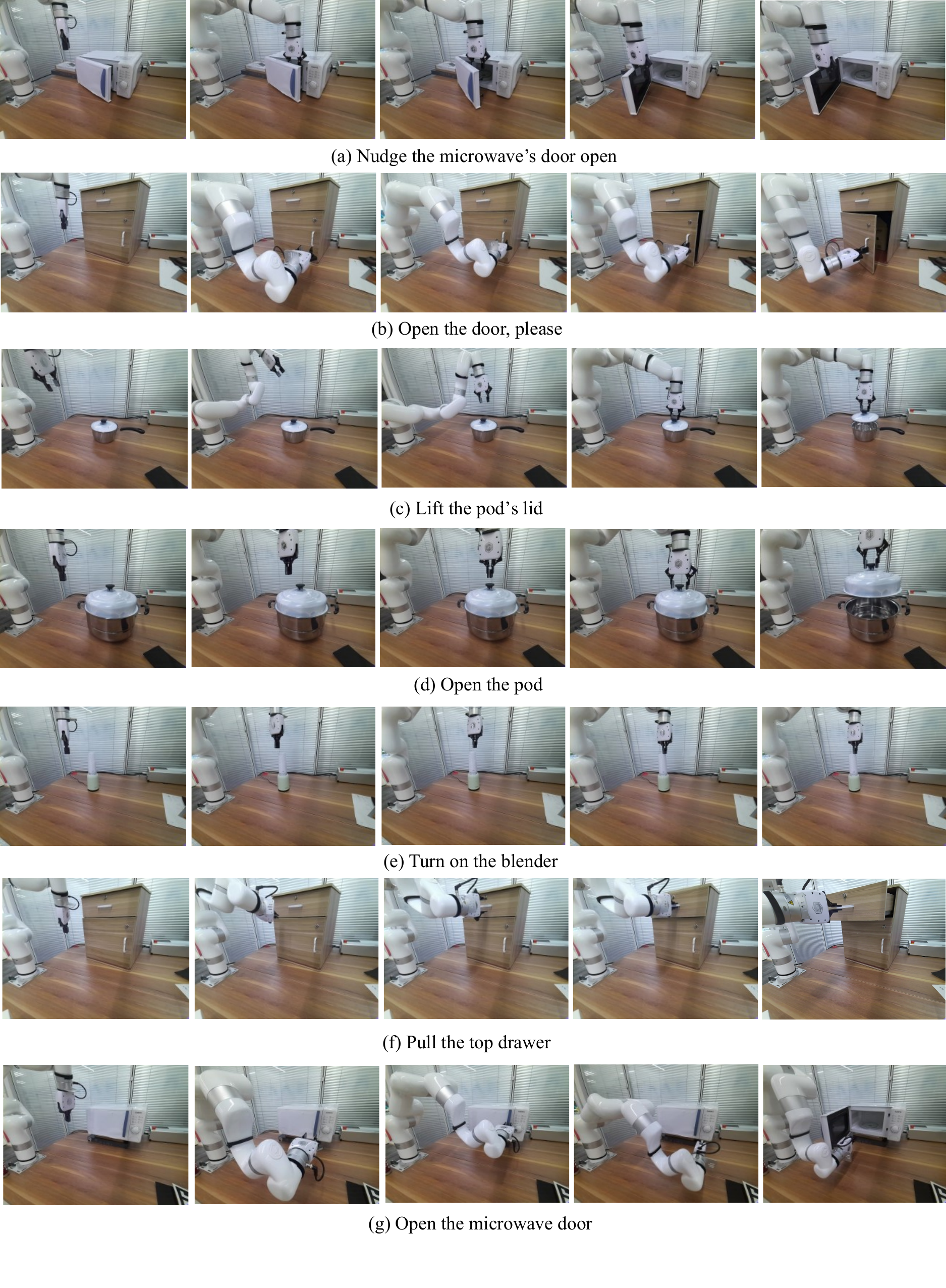}
  \caption{\textbf{Real-robot results.} We show the keyframes of various tasks in our experiments. 
  More results can be found in the \SupplementaryMaterial.}
\end{figure*}

\subsection{Real-Robot Experiments}

\paragraph{Setup}
In our real-world experiments, we establish an experimental setup with the UFACTORY xArm 6 and several different articulated objects for manipulation.

\paragraph{Results}
Fig.~\ref{fig:exp_real_detail} shows our results on real-robot execution, and more results can be found in the \SupplementaryMaterial.
Three challenging cases are highlighted in Fig.~\ref{fig:exp_real_detail} with detailed explanations and intermediate outputs.
The top left is a blender whose top part is perceived as a container for containing juices but functions as a button to be pressed down. Our framework effectively links its semantic and action understandings and successfully executes the task.
The top right shows an emergency stop button for robots, which requires a press (down) to halt an operation and a rotation (up) to restart it. With the auxiliary input of the user manual, our method completes both of these two tasks.
The bottom shows a rather challenging case where the microwave door cannot be directly pulled open. Instead, it requires first pressing the button to initiate a slight door opening. In this case, our method first tries the most straightforward solution of pulling the door and fails. The interactive feedback module then detects this failure and informs the global planner to replan a second strategy that adapts to the current environment, finally completing the task.


\subsection{Generalizable Visual Perception}

\begin{table}[t]
\centering
\resizebox{0.49\textwidth}{!}{
\setlength{\tabcolsep}{1.5pt}
\begin{tabular}{l|c|cc|cc}
\toprule
    MOS $\uparrow$ & GAPartNet 
   & \begin{tabular}[c]{@{}c@{}}VLM\\ (Bard)  \end{tabular}  
   & \begin{tabular}[c]{@{}c@{}}\textbf{Ours}\\ (Bard)  \end{tabular}
   & \begin{tabular}[c]{@{}c@{}}VLM\\ (GPT4V)  \end{tabular}
   & \begin{tabular}[c]{@{}c@{}}\textbf{Ours}\\ (GPT4V)  \end{tabular} \\
\midrule
Part description & 3.8 & 4.1 & 6.3 & 7.2 & \textbf{9.6} \\        
Part accuracy & 8.7 & 4.0 & 6.5 & 6.9 & \textbf{9.5} \\       
Part state precision & 0.0 & 3.7 & 5.6 & 7.7 & \textbf{7.8} \\       
Object \& scene descr. & 0.0 & 2.7 & 6.2 & 6.4 & \textbf{8.0} \\   
Interaction info. & 0.3 & 7.3 & 7.6 & 7.6 & \textbf{8.7} \\      
Overall performance & 2.6 & 4.4 & 6.4 & 7.2 & \textbf{8.7} \\      
\bottomrule
\end{tabular}
}
\caption{\textbf{User study results for scene description.} We solicited volunteers to assess the quality of scene descriptions. Participants were tasked with evaluating the following aspects:
(1) Performance in the overall part description.
(2) Accuracy in the enumeration and naming of parts.
(3) Precision in part-state depiction.
(4) Depiction of objects and the overall scene.
(5) Description of interaction-related information.
(6) Overall performance.}
\label{tab:scene_result}
\end{table}

\begin{figure*}[t]
\begin{minipage}[c]{0.68\linewidth}
\resizebox{\columnwidth}{!}{%
  \begin{tabular}{l|cc|cc|cc|cc}
    \toprule
    Part Perception & \multicolumn{2}{c|}{\textbf{In-distribution}} & \multicolumn{2}{c|}{\textbf{Unseen states}} & \multicolumn{2}{c|}{\textbf{Unseen objects}} & \multicolumn{2}{c}{\textbf{Unseen categories}} \\
    Accuracy& AP@50     & mAP    & AP@50     & mAP    & AP@50              & mAP             & AP@50             & mAP             \\
    \midrule
    PointGroup\citep{jiang2020pointgroup} & 69.70 & 60.58 & 69.54 & 60.48 & 58.26 & 46.29 & 24.57 & 19.40 \\
    SoftGroup\citep{vu2022softgroup} & 69.59 & 60.54 & 70.02 & 59.59 & 59.20 & 47.10 & 28.18 & 22.50 \\
    AutoGPart\citep{liu2022autogpart} & 66.81 & 57.63 & 67.60 & 56.69 & 55.30 & 43.50 & 26.24 & 20.38 \\
    GAPartNet\citep{geng2022gapartnet} & 81.42 & \textbf{72.55} & 80.80 & 71.73 & 63.18 & 53.94 & 36.39 & 27.40 \\ 
    PartGroundedSAM \citep{kirillov2023segany, oquab2023dinov2}   &73.73&61.75&72.36&62.02&66.25&38.64&41.59&28.45\\ 
    \textbf{Ours}
    & \textbf{83.04} & \text{72.23} & \textbf{82.39} & \textbf{71.91} & \textbf{72.17} & \textbf{58.04} & \textbf{47.69} & \textbf{34.57} \\
    \bottomrule
\end{tabular}}
    \captionof{table}{\textbf{Part perception results.} We have curated a novel evaluation dataset for part perception, enriched with more comprehensive data. Our method is benchmarked against 3D point cloud-based techniques, including PointGroup\citep{jiang2020pointgroup}, SoftGroup\citep{vu2022softgroup}, AutoGPart\citep{liu2022autogpart}, and GAPartNet\citep{geng2022gapartnet}. In alignment with the 2D branch of our approach, we also employ SAM\citep{kirillov2023segany} and DINOv2\citep{oquab2023dinov2} to establish a 2D-centric baseline. For evaluation, we adopt AP@50 and mAP as our primary metrics.}
    \label{tab:perception_results}
\end{minipage}
\hfill %
\begin{minipage}[c]{0.3\linewidth}
    \centering
    \includegraphics[width=0.98\linewidth, trim=0 .5cm 0 .5cm, clip]{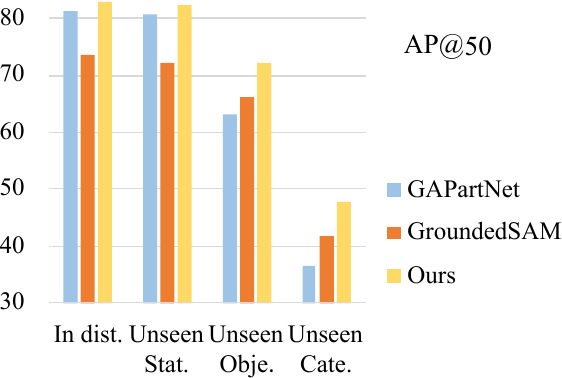}
    \caption{\textbf{Part perception results.}
    Our method consistently achieved the highest AP@50 score for each level of generalization.}
    \label{fig:3dcompare}
  \end{minipage}
\end{figure*}

As described in Sec. \ref{sec:method}, we design knowledge fusion mechanisms to join the force VLMs and small domain-specific models both in our context comprehension and part perception. In this section, we evaluate our intermediate results on scene description (Sec. \ref{sec:method:scene_percep}) and part perception (Sec. \ref{sec:grounding}) and show that our generalizable perception modules achieve a better balance between generalization and exactness compared to existing methods or its ablated versions.

\paragraph{Scene description (Sec. \ref{sec:method:scene_percep})}
In our instruction interpreter, a scene description is generated from the visual observations to inform the other modules of the scene context.
To evaluate the quality of scene descriptions generated by our method which joins the forces of generalist and specialist models, 
we curate an image dataset specifically for evaluating scene descriptions for manipulation purposes. The dataset includes 145 images from 15 object categories with varying part states.
We also designed six interaction-oriented metrics for assessing the qualities of the descriptions: part description, part accuracy, part state precision, object\&scene description, interaction-related information, and overall performance.
With these metrics, we can collect the Mean Opinion Score (MOS) through carefully designed user studies, as in \citep{gal2022image}.

We evaluate the scene descriptions generated by five methods:  (1) GAPart Model: detection results are translated into natural language descriptions through a human-crafted format; (2)(3) VLM (Bard, GPT-4V), craft descriptions from the input image using Bard and GPT-4V; (4)(5) Our methods (based on Bard and GPT-4V).
In the user study, participants are provided an image and the descriptions generated by the three methods at each time, and they are asked to rate the descriptions using a scale of 0-10 (with 10 being the best) according to the 6 metrics. 17 participants have been invited, and each person is asked to rate about 40 scenes on average.

Table~\ref{tab:scene_result} shows the results of this user study. We found that domain-specific models can hardly provide useful object and scene-level descriptions due to their lack of general knowledge. On the other hand, VLM provides rich context but lacks accuracy in domain-related information. Our method benefits from both of them and achieves much better performances by joining their strengths.



\paragraph{Part perception (Sec. \ref{sec:grounding})}
In our part perception task, we utilize a single RGBD image as input to predict part-related information, including part semantic segmentation, pose and state estimation. To evaluate the methods, we introduce a new benchmark for part perception tasks. In comparison with GAPartNet, our benchmark is more comprehensive and suitable for manipulation tasks. For instance, we incorporate a greater number of objects with closed parts, accounting for 12.5\%, which is not included in GAPartNet\citep{geng2022gapartnet}.
To enable the evaluation of generalizations at different levels, the test data are divided into 4 subsets: in-distribution, unseen (articulation) states, unseen objects, and unseen object categories.
We use the average precision AP@50 and mAP as evaluation metrics for part segmentation, which are widely adopted in prior 3D semantic/instance segmentation benchmarks such as ScanNet\citep{dai2017scannet} and GAPartNet\citep{geng2022gapartnet},

Table \ref{tab:perception_results} presents the results of our method in comparison to various baselines for part perception. We consider GAPartNet\citep{geng2022gapartnet}, a modified version of PointGroup\citep{jiang2020pointgroup}, SoftGroup\citep{vu2022softgroup}, and AutoGPart \citep{liu2022autogpart} as our primary baselines. Additionally, we adapted GroundedSAM, rebranding it as PartGroundedSAM for this context. The performance metrics in Table \ref{tab:perception_results} indicate that our approach surpasses other baselines. Notably, we observed that methods based on 3D tend to under-perform on out-of-domain data, whereas 2D-centric methods yield subpar results for parts. Our methodology derives advantages from both the 2D and 3D realms, resulting in optimal performance and superior generalization capabilities.

\section{Conclusions}
In this paper, we introduce a novel framework for language-guided manipulation of articulated objects.
Bridging the understanding of object semantics and actionability at the part level, we can ground language-implied actions to executable manipulations.
Throughout our framework, we also study the combination of general-purpose large vision/language models and domain-specialist models for enhancing the richness and correctness of network predictions.  better handle these tasks and achieve state-of-the-art performance.
We demonstrate our strong generalizability across diverse object categories and tasks. We also provide a new benchmark for language-instructed articulated-object manipulations.

\paragraph{Limitations and future work}  
Since the SAGE system is built upon existing large vision-language models, it suffers from the drawbacks of demanding massive data and massive computing power. Although in our case we use APIs provided by GPT-4V, the long inference time and high cost could be a concern. On the other hand, we rely on the expert model GAPartNet to detect parts and its prior knowledge to manipulate the parts. The policy is not always optimal by blindly following the pre-defined end-effector trajectories. Such a motion-planning-based execution policy is not as responsive as a reinforcement learning or imitation learning agent. One future exploration direction could be fine-tuning existing large models to directly output the desired low-level action of the end-effector to increase responsiveness.



\bibliographystyle{plainnat}
\bibliography{references}

\end{document}


\maketitle


\section{More details for the Benchmark}


\clearpage
\acknowledgments{If a paper is accepted, the final camera-ready version will (and probably should) include acknowledgments. All acknowledgments go at the end of the paper, including thanks to reviewers who gave useful comments, to colleagues who contributed to the ideas, and to funding agencies and corporate sponsors that provided financial support.}


\bibliography{example}  


\title{\LARGE Supplementary Materials for \\
\textbf{SAGE}: Bridging \textbf{{S}}emantic and \textbf{{A}}ctionable Parts for
\textbf{{Ge}}neralizable Manipulation of Articulated Objects}

\author{Author Names Omitted for Anonymous Review. Paper-ID 70}



%


\twocolumn[{%
\renewcommand\twocolumn[1][]{#1}%
\maketitle
}]


\section{Supplementary Video}
Please refer to the accompanying video for a comprehensive and intuitive overview of our methodology. The video includes demonstrations from real-world experiments with two interesting examples, offering a thorough analysis and detailed walk-through of our pipeline. We highly recommend watching the video for a clearer understanding of our work.

\section{Part-aware Scene Perception}

\subsection{Data sources.} Our dataset is primarily generated through the rendering of articulated data assets sourced from the GAPartNet dataset \citep{geng2022gapartnet}, itself an amalgamation of two pre-existing datasets: PartNet-Mobility \citep{xiang2020sapien} and AKB-48 \citep{liu2022akb}. From GAPartNet, we curated a selection of 27 object categories for our analysis.

\subsection{Dataset Rendering}

In line with the methodologies outlined in \citep{geng2022gapartnet}, our approach leverages the SAPIEN 2.0 platform \citep{xiang2020sapien} for the generation of a comprehensive dataset from our meticulously selected objects. This dataset encompasses partial point clouds, part semantic and instance segmentation masks, Normalized Point Coordinate Space (NPCS) maps, and part pose annotations. These components collectively furnish the requisite data for nuanced part perception and grounding tasks.

\textbf{Environment Settings.} To enhance the realism of our rendered scenes, we activate the ray-tracing feature in SAPIEN. The process involves varying the articulation points of objects and selecting camera angles from pragmatically determined perspectives. For each object category, we specifically tailor the camera positioning range to ensure optimal visibility, avoiding perspectives that obscure significant details, such as the rear of StorageFurniture or the underside of an Oven, and maintaining a balanced distance. Concurrently, we modulate the ambient lighting between 10\% and 90\% intensity and permit slight camera rotations of up to $\pm 5^\circ$ to introduce a natural variability in our dataset's visual representation.
The output image resolution is set to $800 \times 800$. For each object, we render 32 RGB images. Along with each RGB image, we also obtain the segmentation masks and the depth image using the built-in features of the SAPIEN environment. Additionally, we compute NPCS maps and oriented tight bounding boxes as part pose annotations for all GAParts.
\textbf{Enhanced Balance.} In a departure from the practices of GAPartNet \citep{geng2022gapartnet}, we recognized the significance of representing objects in their closed state for tasks involving articulated object manipulation. To address this, we enriched our dataset by incorporating an additional 30

\textbf{Point Cloud Generation.} We employ back-projection techniques using camera intrinsics alongside 2D RGB and depth images to generate dense, partial point clouds. Each of these dense clouds is then sampled down to 20,000 points through Farthest-Point-Sampling (FPS) to maintain uniformity and manageability. During this sampling process, we concurrently produce the corresponding ground truth for semantic and instance segmentation, as well as NPCS maps. These processed point clouds and their annotations are prepared offline to expedite subsequent 3D processing tasks.

\subsection{Training Procedure}
We adopt an end-to-end training approach, setting the maximum number of epochs at 200. The training utilizes the Adam optimizer, with a batch size of 32 and an initial learning rate set to 0.001. This training regimen is executed over approximately 1.5 days using a single NVIDIA GeForce RTX 4090 GPU. Furthermore, to enhance model performance, we progressively apply a multi-resolution training strategy, which has been shown to significantly improve results.




\subsection{Simulation Experiments}

\textbf{Benchmark Settings.} We set up our interaction environment using the SAPIEN\citep{xiang2020sapien} simulator, modified from the ManiSkill challenge\citep{mu2021maniskill}. We benchmark our method on several tasks. These tasks exemplify robot manipulation under the motion constraint of a prismatic or a revolute joint. For evaluation, we randomly pick unseen objects that contain target semantics with target parts from seen and unseen object categories. Considering the limitation of the single gripper, we select such objects that, given the ground truth of their segmentation and pose, can be opened successfully using the heuristics under our benchmark setting.  Compared to the ManiSkill Challenge\citep{mu2021maniskill}, we provide diverse and challenging language instructions and limit our observation to a first-frame-only partial point cloud of the object, with only one point around the part center indicating which part to interact with. Given the initial state of the robot, it performs the whole manipulation only based on the observation at the first step until the global planner has new action units. The action space of the robot is the motor command of the 6 joints of the robot to determine the pose of the gripper, and we use position control to open or close grippers. A success is defined as manipulating the part for 90\% of the motion range within 1,000 steps with a stable stop at the end.

\textbf{Part-pose-based Manipulation Heuristics.}
Following GAPartNet\citep{geng2022gapartnet}, we use the interaction policy based on the heuristics to open drawers, open doors, manipulate handles, and press buttons and so on. Specifically, when we get the part pose, we can immediately get the grasping pose with our policy. Then we use a motion planning library (\ie, mplib, provided by SAPIEN \citep{xiang2020sapien}), to move our gripper to the grasping pose. Then, with our interaction policy and axis predicted from our method, we design the end-effector trajectory just along the trajectory of the part moving and interpolate the trajectory with a time step of $\frac{1}{250}$. With the IK (Inverse Kinematics) algorithm and a PID controller, we solve the poses of joints and move the end-effector along the trajectory. All of our implementations are decoupled from ROS and can be easily implemented in other simulators.

\subsection{Real-world Experiments}

\textbf{Implementation Details.}
To evaluate the robustness and generalizability of our method, we use XArm to manipulate previously unseen objects with only partial point cloud observations in the real world. We use similar motion planning and a similar end-effector trajectory as what we do in the simulator. A partial point cloud of the target object is acquired from the RGB-D camera (Kinect DK sensor in our experiments). To set up the interaction environment, we place the object and the robot arm in a proper position for interaction and use ArUco markers to calibrate the camera sensor. We also provide a point indicating the part to interact with, just like in the simulator. During manipulation, we first estimate the bounding box of the target part and calculate the trajectory using the heuristics, then use the control API provided by the robot arm to follow the trajectory and finish the task. Overall, we conduct several manipulation tasks with challenging language instructions.

\section{More Experiment Results}
\begin{figure*}
    \centering
    \includegraphics[width=.95\linewidth]{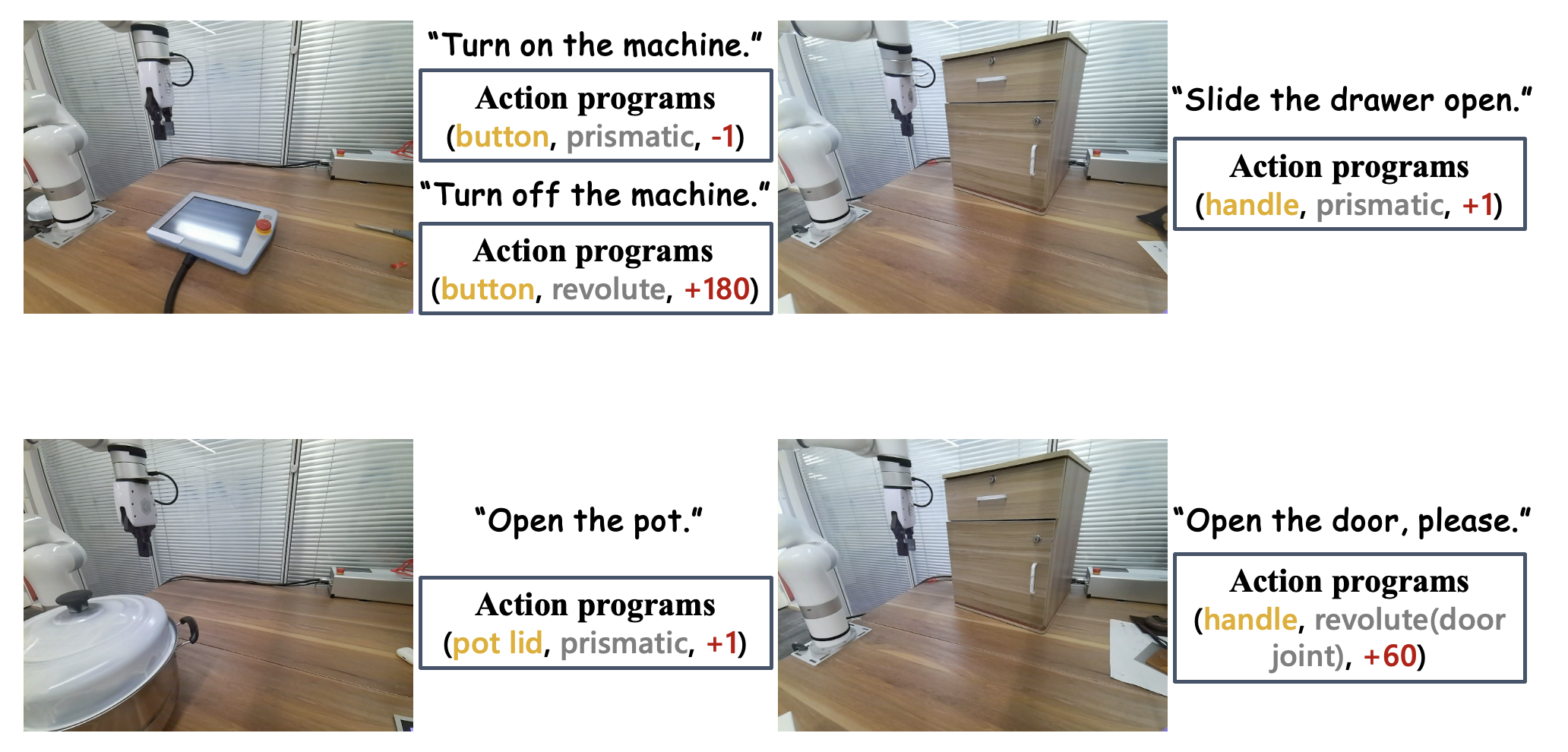}
    \caption{\textbf{More Real-world Results.} More real-world experiment results and corresponding action programs.}
    \label{fig:supp}
\end{figure*}

\begin{figure*}
    \centering
    \includegraphics[width=.95\linewidth]{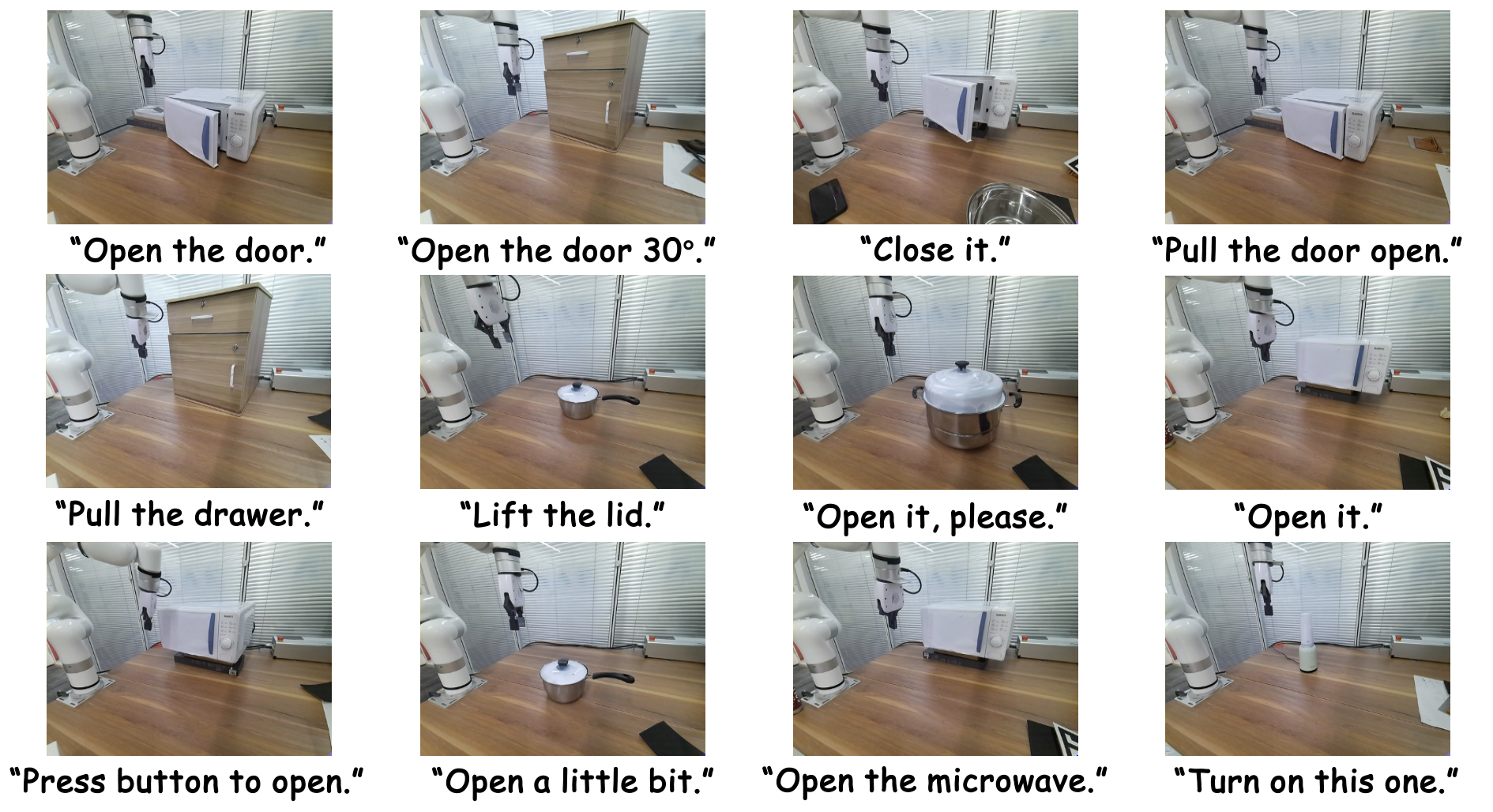}
    \caption{\textbf{More Real-world Results.} More real-world experiment results with diverse instructions.}
    \label{fig:supp2}
\end{figure*}

We provide more real-world experiment results in Fig.\ref{fig:supp} and Fig.\ref{fig:supp2}. In Fig.\ref{fig:supp}, we provide more real-world experiment results and corresponding action programs. In 
Fig.\ref{fig:supp2}, we provide more real-world experiment results with diverse instructions.

\clearpage


\bibliographystyle{plainnat}
\bibliography{references}